\documentclass[11pt,a4paper]{article}
\usepackage[utf8]{inputenc}
\usepackage[T1]{fontenc}
\usepackage[margin=0.86in]{geometry}
\usepackage{lmodern}
\usepackage{microtype}
\usepackage{amsmath,amssymb,mathtools}
\usepackage{graphicx}
\usepackage{booktabs,multirow,array}
\usepackage[section]{placeins}
\usepackage[numbers,sort&compress]{natbib}
\usepackage{xcolor}
\usepackage{enumitem}
\usepackage{hyperref}

\hypersetup{
    colorlinks=true,
    linkcolor=blue!50!black,
    citecolor=teal!55!black,
    urlcolor=blue!50!black,
    pdftitle={MolGPAO: Molecular Generative Pre-trained Autoregressive Octrees with Seamless Speculative Decoding},
    pdfauthor={Shangqian Chen, Guojiang Zhao, Zheng Cheng, Jiankun Wang, Hongshuai Wang, Zhifeng Gao, Linfeng Zhang}
}

\title{\textbf{MolGPAO: Molecular Generative Pre-trained Autoregressive Octrees with Seamless Speculative Decoding}}

\author{
Shangqian Chen, Guojiang Zhao, Zheng Cheng, Jiankun Wang,\\
Hongshuai Wang, Zhifeng Gao, Linfeng Zhang\\[4pt]
\small DP Technology, Beijing, China\\
\footnotesize\texttt{\{chenshangqian, zhaogj, chengzheng, wangjiankun, wanghongshuai, gaozf, zhanglf\}@dp.tech}
}

\date{November 2026}

\begin{document}
\maketitle

\begin{abstract}
Autoregressive generation over adaptive three-dimensional octrees offers an intrinsic $\mathcal{O}(N \log N)$ spatial sparsity advantage over dense $\mathcal{O}(N^2)$ diffusion and flow-matching models on large drug-like molecules. In practice, however, unbonded octree autoregressive models have trailed explicit-bond flow-matching architectures in equilibrium geometry, closed-shell valence stability, and conditional property fidelity due to octant-seam coordinate jumps, leaf-dominated compute allocation, and target-shortcut collapse in internal process critics. Here we introduce \textbf{MolGPAO}, a standalone 3D octree autoregressive architecture with seamless speculative decoding that produces a single molecule per prompt using zero external surrogate predictors, zero reinforcement learning, and zero inference-time interatomic potential relaxation. \textbf{MolGPAO} combines scale-free atom-budget modulation and hierarchical packing priors, a vectorized pause-and-gather speculative octree funnel governed by a continuous combinatorial complexity law, an anti-shortcut equivariant multipole process critic supervised solely on dataset ground-truth labels, a sub-millisecond $SE(3)$ seam-healing refiner paired with zero-displacement valence reconciliation, and synchronization-free GPU leaf decoding. Across unconditional drug-scale generation on \texttt{GEOM-Drugs}, six-property conditional generation on \texttt{QM9}, and frontier-orbital inverse design on \texttt{OE62}, single-output \textbf{MolGPAO} lifts drug-scale valid-connected yield $6.78\times$ to $86.2\%$, decodes $100$-atom structures up to $7.3\times$ faster than $100$-step \texttt{FlowMol}, surpasses six-model \texttt{PropMolFlow} on \texttt{QM9} frontier orbital and dipole benchmarks at $362\times$ lower training compute, and doubles live \texttt{PBE0/def2-TZVP} quantum hit yield on \texttt{OE62} at $92.8\times$ lower training compute.
\end{abstract}

\section{Introduction}
\label{sec:intro}

De novo generation of three-dimensional (3D) molecular structures underpins structure-based drug discovery \citep{axelrod2022geom,dunn2024flowmol,irwin2025semlaflow} and functional organic optoelectronic design \citep{stuke2020oe62,zhao2025virtual,zhao2026o2gen,wang2026mimflow,zeng2026propmolflow}. Contemporary 3D molecular generative architectures divide into two paradigms with contrasting computational and physical trade-offs:

\begin{enumerate}[leftmargin=1.6em,itemsep=2pt,topsep=3pt]
    \item \textbf{Dense Iterative Equivariant Diffusion and Flow Matching.} Coordinate-only diffusion models such as \texttt{EDM} \citep{hoogeboom2022edm} and \texttt{GeoLDM} \citep{xu2023geoldm} integrate $500\text{--}1{,}000$ $E(3)$-equivariant message-passing steps over complete graphs and suffer low molecule-level valence stability on drug-scale molecules ($0.2\%\text{--}5.2\%$ on \texttt{GEOM-Drugs}). Joint 2D+3D diffusion and continuous-time Markov chain (CTMC) flow-matching models---\texttt{MiDi} \citep{vignac2023midi}, \texttt{EQGAT-diff} \citep{le2023eqgatdiff}, \texttt{FlowMol} \citep{dunn2024flowmol}, \texttt{SemlaFlow} \citep{irwin2025semlaflow}, and \texttt{PropMolFlow} \citep{zeng2026propmolflow}---explicitly co-evolve all $N(N-1)/2$ pairwise bond states alongside 3D coordinates across $100\text{--}500$ steps. While explicit edge evolution restores chemical validity, materializing dense $\mathcal{O}(N^2)$ pair tensors at every step imposes steep latency growth from $N=30$ to $N=100$ atoms and requires $52\text{--}156\,\text{GPU-h}$ of training compute \citep{zeng2026propmolflow}.
    \item \textbf{Sparse Hierarchical 3D Octree Autoregression.} \texttt{Uni-3DAR} \citep{lu2025uni3dar} recursively subdivides 3D space from a cubic root ($L_{\max} \in \{6, 7\}$, box side $15.36\text{--}30.72\,\text{\AA}$) down to $0.24\,\text{\AA}$ leaf voxels ($\ell = 0$) in breadth-first Morton order, emitting only non-empty spatial octants ($\mathcal{O}(N \log N)$ tokens) followed by chemical element $Z_i$ and quantized $0.01\,\text{\AA}$ sub-voxel offsets $(\Delta x_i, \Delta y_i, \Delta z_i)$ via single-pass causal KV-cached decoding.
\end{enumerate}

Despite its $\mathcal{O}(N \log N)$ complexity, deploying vanilla octree autoregression on drug-scale molecules (\texttt{GEOM-Drugs}) and conditional quantum design (\texttt{QM9}, \texttt{OE62}) without external property surrogates or inference-time MLIP relaxation exposes three foundational bottlenecks:

\begin{itemize}[leftmargin=1.5em,itemsep=2pt,topsep=3pt]
    \item \textbf{Morton-Seam Coordinate Jumps and Valence Fragility.} When two covalently bonded atoms straddle an octree division plane, their sub-voxel coordinates are decoded far apart in the causal Morton sequence without explicit pairwise bond springs. Cross-seam misalignments of $0.04\text{--}0.08\,\text{\AA}$ trigger zero-step atomic force spikes ($90$th-percentile $F_{\max} > 17\text{--}22\,\text{eV/\AA}$ under \texttt{DPA-4} \citep{deepmodeling2026dpa4}) and cause post-hoc bond perception (\texttt{RDKit DetermineBonds} \citep{rdkit2026bonds}) to assign spurious zwitterions on $12\%\text{--}32\%$ of structures.
    \item \textbf{Leaf-Heavy Compute Allocation and Integer-Count Trapping.} Coarse octree splits ($\ell \ge 1$) consume $<30\%$ of autoregressive FLOPs, whereas $\ell = 0$ leaf-atom and $0.01\,\text{\AA}$ coordinate heads consume $>70\%$. Naive leaf-complete sampling wastes over $70\%$ of GPU compute finishing strained or off-target coarse envelopes, while rigid integer atom quotas per octant starve large scaffolds ($N \ge 45$).
    \item \textbf{Target-Shortcut Collapse in Self-Contained Critics and Sync Stalls.} Pruning octree branches internally without external predictors requires inferring quantum properties directly from partial octrees and 3D point clouds supervised solely on dataset ground-truth labels. Naive supervision on teacher-forced pairs $(\mathcal{M}_i, t_i = y_i)$ collapses onto prompt target $t_i$, shifts hidden states under classifier-free guidance (CFG) \citep{ho2022cfg}, and blurs covalent bond orders under coarse radial bins, while Python-driven octree queues incur $>200$ GPU--CPU synchronization stalls per batch.
\end{itemize}

\begin{figure}[!t]
  \centering
  \includegraphics[width=\textwidth]{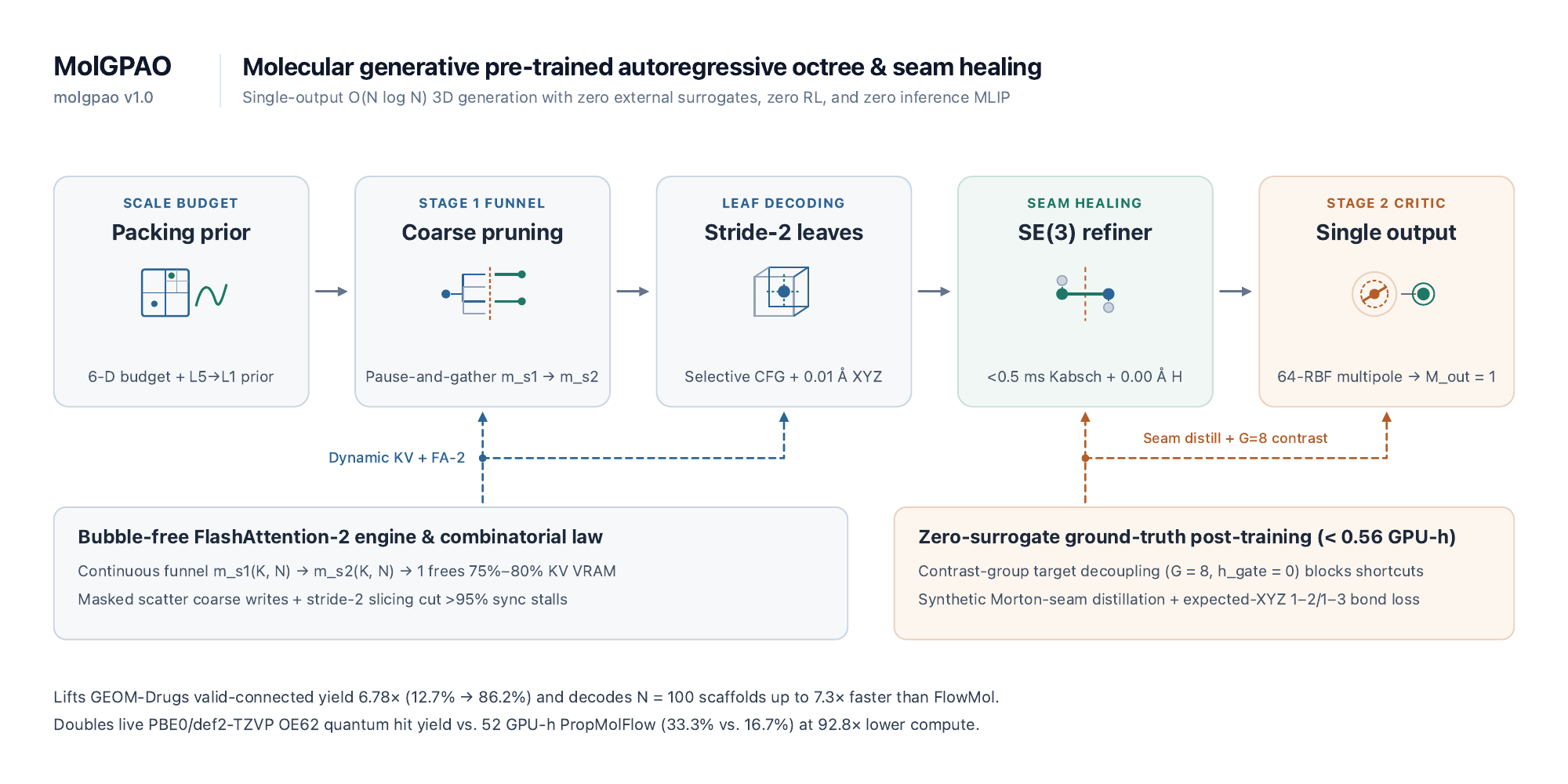}
  \caption{\textbf{Overview of the MolGPAO architecture and seamless speculative octree decoding.} Scale-free budget modulation and $\text{L5}\to\text{L1}$ packing priors feed Stage-1 pause-and-gather coarse octree pruning ($m_{\text{s1}}\to m_{\text{s2}}$), synchronization-free stride-$2$ leaf decoding under selective \texttt{CFG}, sub-millisecond $SE(3)$ \texttt{PostTreeGraphRefiner} Morton-seam healing with zero-displacement H reconciliation, and an anti-shortcut $64$-RBF multipole Stage-2 critic emitting one molecule per prompt ($M_{\text{out}} \equiv 1$).}
  \label{fig:arch}
\end{figure}

Here we present \textbf{MolGPAO} (\emph{Molecular Generative Pre-trained Autoregressive Octrees}), a standalone octree architecture with seamless speculative decoding that produces one molecule per prompt ($M_{\text{out}} \equiv 1$) using zero external surrogate models, zero reinforcement learning, and zero training- or inference-time MLIP calls (Figure~\ref{fig:arch}). Under full $\text{SO}(3)$ equivariant augmentation and zero $N_{\text{atoms}}$ leakage into intensive property conditioning, \textbf{MolGPAO} introduces five synergistic mechanisms:
\begin{enumerate}[leftmargin=1.6em,itemsep=1pt,topsep=2pt]
    \item \textbf{Scale-Free Budget Modulation and Packing Priors (Section~\ref{sec:budget}):} A zero-initialized 6-D continuous scale embedding paired with empirical $\text{L5}\to\text{L1}$ organic packing priors that expand $N=16\text{--}140$ scaffolds smoothly without octant count trapping;
    \item \textbf{Speculative Octree Decoding (Section~\ref{sec:spectree}):} A vectorized FlashAttention-2 \citep{dao2023flashattention2} pause-and-gather KV funnel $m_{\text{s1}}(K,\hat{N}) \to m_{\text{s2}}(K,\hat{N}) \to 1$ governed by a continuous size-complexity law across search tiers $K \in \{1, 2, 4, 8, 12, 16\}$;
    \item \textbf{Equivariant Multipole Process Critic (Section~\ref{sec:critic}):} Pure 3D geometric contrast-group supervision ($p_{\text{decouple}}=0.50$), $64$-RBF covalent continuous-filter edge gating, and $\text{SO}(3)$-invariant 12-channel equivariant multipole norms trained solely on dataset ground-truth labels;
    \item \textbf{Octant-Seam Healing and Valence Grounding (Section~\ref{sec:refiner}):} A sub-millisecond $4$-layer $SE(3)$ \texttt{PostTreeGraphRefiner} distilled on synthetic Morton-seam perturbations prior to Stage-2 critic scoring, paired with expected-\texttt{XYZ} bonded-distance SFT, selective logit-CFG, and zero-displacement valence reconciliation;
    \item \textbf{Batched GPU Decoding and Memory Scheduling (Section~\ref{sec:engine}):} Stride-$2$ leaf slicing and budget-proportional KV allocation that eliminate $>95\%$ of host-device synchronization stalls.
\end{enumerate}

\section{MolGPAO Architecture}
\label{sec:method}

\subsection{Scale-Free Budget Modulation and Packing Priors}
\label{sec:budget}

Let an octree of root merge level $L_{\max} \in \{6, 7\}$ ($0.24\,\text{\AA}$ leaf voxel width) represent a 3D molecule $\mathcal{M} = \{(Z_i, \mathbf{r}_i)\}_{i=1}^N$. Standard integer root-count prediction commits early coarse octants to rigid atom quotas, causing steric collisions on small molecules or premature termination on $N \ge 50$ scaffolds, whereas concatenating $N$ into a property conditioner conflates intensive electronic observables with extensive molecular size.

\textbf{MolGPAO} decouples extensive size control from property conditioning via two mechanisms:
\begin{enumerate}[leftmargin=1.6em,itemsep=2pt,topsep=2pt]
    \item \textbf{Scale-Free Budget Embedding:} Every non-EOT octree token at level $\ell \in \{0, \dots, L_{\max}\}$ receives a zero-initialized residual embedding $\phi_{\text{budget}}(\ell, N) \in \mathbb{R}^d$ computed from dimensionless log-scale and harmonic features of target atom budget $N$:
    \begin{equation}
    \label{eq:budget_emb}
    \begin{aligned}
    \mathbf{z}_{\text{budget}}(\ell, N) = \Big[&\mathbb{I}_{N>0},\; \frac{\log(1+N)}{4.5},\; \operatorname{clip}\!\left(\frac{N-30}{25}, -2.5, 4.5\right), \\
    &\frac{\ell+1}{8},\; \sin\!\left(\pi \log\frac{N}{30}\right),\; \cos\!\left(\pi \log\frac{N}{30}\right)\Big],
    \end{aligned}
    \end{equation}
    trained with $p_{\text{drop}} = 0.12$ budget dropout so one checkpoint supports both exact-budget and unbudgeted free generation across $N \in [16, 140]$.
    \item \textbf{Hierarchical Packing Priors ($\text{L5}\to\text{L1}$):} During coarse expansion ($1 \le \ell \le 5$), let $R_\ell$ denote remaining unexpanded cells at level $\ell$ and $C_{\ell-1}$ child cells already spawned at level $\ell-1$. Equilibrium organic conformers follow characteristic occupancy ratios $\delta_\ell = N / N_{\text{cells}}(\ell) \approx (3.00, 1.36, 0.99, 0.99, 0.99)$ from $\ell=5$ ($7.68\,\text{\AA}$) down to $\ell=1$ ($0.48\,\text{\AA}$). For any 8-bit split mask $u \in \{1,\dots,255\}$ with popcount $k(u) = \|u\|_1$, we bias cardinality while preserving all $\binom{8}{k}$ spatial orientations:
    \begin{equation}
    \label{eq:vol_prior}
    \tilde{\ell}_{\text{tree}}(u) = \ell_{\text{tree}}(u) - w_\ell \left(k(u) - \operatorname{clip}\!\left(\frac{N/\delta_\ell - C_{\ell-1}}{R_\ell},\, 1,\, \kappa_\ell\right)\right)^2 + \mathcal{P}_{\text{steric}}(u; \ell),
    \end{equation}
    applying soft quadratic guidance on coarse envelopes ($\ell \in \{3,4,5\}$) and steric leaf caps ($k(u) \le 2$) on fine $0.96/0.48\,\text{\AA}$ cells ($\ell \in \{1,2\}$).
\end{enumerate}

\subsection{Speculative Octree Decoding}
\label{sec:spectree}

\paragraph{Continuous Combinatorial Branching Law ($M_{\text{out}} \equiv 1$).}
To compare directly against single-shot diffusion and flow-matching models without external best-of-$M$ re-ranking, \textbf{MolGPAO} outputs strictly one molecule per prompt ($M_{\text{out}} \equiv 1$) across internal search tiers $K \in \{1, 2, 4, 8, 12, 16\}$. Compact scaffolds ($N \le 20$) saturate coarse topological diversity rapidly, whereas larger scaffolds ($N \ge 30$) require wider prefix exploration. We therefore define a smooth complexity factor $\eta(\hat{N}) = \operatorname{clip}((\hat{N} - 16)/14,\, 0,\, 1)$ ($\eta = 1$ when unbudgeted) over median batch atom budget $\hat{N}$:
\begin{equation}
\label{eq:funnel_law}
m_{\text{s1}}(K, \hat{N}) = 2 \left\lceil \frac{K \big(4 + 4\,\eta(\hat{N})\big)}{2} \right\rceil, \;
m_{\text{s2}}(K, \hat{N}) = 2 \left\lceil \frac{K \big(1.25 + 0.75\,\eta(\hat{N})\big)}{2} \right\rceil, \;
M_{\text{out}} \equiv 1 \; (K > 1),
\end{equation}
and $(1 \to 1 \to 1)$ for $K=1$. Equation~\eqref{eq:funnel_law} scales continuously from compact \texttt{QM9} prompts ($\hat{N} \approx 18$, giving funnels $10\to 4\to 1$ through $74\to 22\to 1$ across $K=2\text{--}16$) to drug and semiconductor scaffolds ($\hat{N} \ge 30$, giving $16\to 4\to 1$ through $96\to 24\to 1$ across $K=2\text{--}12$).

\paragraph{Pause-and-Gather Funnel.}
For a batch of $B_p$ prompts, two internal GPU pruning stages execute sequentially:
\begin{enumerate}[leftmargin=1.6em,itemsep=2pt,topsep=2pt]
    \item \textbf{Stage 1 (Coarse Envelope Pruning, $B_p m_{\text{s1}} \to B_p m_{\text{s2}}$):} All $B_p m_{\text{s1}}$ trajectories decode coarse splits ($\ell \ge 1$) in parallel and pause upon reaching their first $\ell = 0$ leaf token. \texttt{OctreeProcessCritic} combines pooled coarse Transformer states with $14$ $\text{SO}(3)$-invariant voxel-envelope descriptors (leaf count, radius of gyration, principal inertia eigenvalues, anisotropy and planarity ratios, nearest-neighbor packing, and coarse split perplexity) to retain the top-$m_{\text{s2}}$ prefixes per prompt via vectorized KV gathering, freeing $75\%\text{--}80\%$ of KV memory prior to leaf decoding.
    \item \textbf{Stage 2 (Seam-Healed Leaf Pruning, $B_p m_{\text{s2}} \to B_p$):} Surviving prefixes complete $\ell = 0$ leaf atoms and sub-voxel offsets, undergo sub-millisecond \texttt{PostTreeGraphRefiner} seam healing, and are ranked by \texttt{OctreeElementProcessCritic} to emit a single molecule ($M_{\text{out}} = 1$) per prompt.
\end{enumerate}

\subsection{Equivariant Multipole Process Critic}
\label{sec:critic}

Supervising an internal critic naively on teacher-forced tuples $(\mathcal{M}_i, t_i = y_i)$ using conditioned Transformer states $\mathbf{h}$ induces trivial shortcut collapse ($\hat{y} \approx t_i$) under test-time CFG rollouts. \textbf{MolGPAO} eliminates shortcut collapse using only dataset ground-truth labels via three designs:

\begin{enumerate}[leftmargin=1.6em,itemsep=2pt,topsep=2pt]
    \item \textbf{Pure Geometric Readout and Contrast-Group Supervision:} Setting \texttt{h\_gate\_scale}$=0$ detaches Stage-2 \texttt{OctreeElementProcessCritic} completely from conditioned Transformer states, preventing teacher-forced versus CFG distribution shift. Each ground-truth conformer $\mathcal{M}_i$ ($\pm 0.015\,\text{\AA}$ jitter) is paired with matched ($t = y_i$) and mismatched ($t' \ne y_i$, $p_{\text{decouple}}=0.50$) targets in contrast groups of size $G=8$, mixed with unlabeled on-policy rollouts that supervise chemical legality and coordination validity. Within each group, the critic optimizes Huber property regression alongside ListNet KL and pairwise ranking over composite utility $U_g = -\|\hat{\mathbf{y}}_g - \mathbf{t}_g\|_1/\sigma_y + \log p_{\text{legal}} + \log p_{\text{constr}}$.
    \item \textbf{High-Resolution Covalent Edge Gating:} Three layers of continuous-filter message passing over $64$ narrow Gaussian radial basis functions spanning $d_{ij} \in [0.65, 5.20]\,\text{\AA}$ ($\gamma = 42.0\,\text{\AA}^{-2}$, $\sigma \approx 0.11\,\text{\AA}$) with smooth cosine cutoff and element-pair value gates resolve single ($1.54\,\text{\AA}$), aromatic ($1.39\,\text{\AA}$), double ($1.34\,\text{\AA}$), and triple ($1.20\,\text{\AA}$) bonds alongside $36$ stoichiometric and coordination features.
    \item \textbf{$\text{SO}(3)$-Invariant Multipole Norms:} Because scalar node pooling cannot detect dipole cancellation in centrosymmetric isomers, a linear head on final scalar node states $\mathbf{h}_i^{(3)}$ predicts $C=12$ charge-neutralized channels $\tilde{q}_{i,c} = q_{i,c} - \frac{1}{N}\sum_{j=1}^N q_{j,c}$ analogous to partial-charge electrostatic parametrizations in conjugated molecular force fields \citep{zhao2025oscff} over centered coordinates $\tilde{\mathbf{r}}_i = \mathbf{r}_i - \bar{\mathbf{r}}$, yielding exact $\text{SO}(3)$-invariant dipole norms and quadrupole traces:
    \begin{equation}
    \label{eq:multipole}
    \mu_c(\mathcal{M}) = \left\| \sum_{i=1}^N \tilde{q}_{i,c}\, \tilde{\mathbf{r}}_i \right\|_2, \qquad
    Q_c(\mathcal{M}) = \sum_{i=1}^N \tilde{q}_{i,c}\, \|\tilde{\mathbf{r}}_i\|_2^2, \qquad c \in \{1,\dots,12\}.
    \end{equation}
    Concatenating $\{\mu_c, Q_c\}_{c=1}^{12}$ with sum and mean node readouts achieves blind mismatched-target ($t' \ne y$) dev Pearson correlations of $r = 0.9867\text{--}0.9989$ across all six \texttt{QM9} quantum properties with zero external predictors.
\end{enumerate}

\subsection{Octant-Seam Healing and Valence Grounding}
\label{sec:refiner}

\paragraph{$SE(3)$-Equivariant Seam Refiner.}
To heal Morton octant-seam discontinuities without MLIP calls, a $4$-layer $SE(3)$-equivariant graph transformer (\texttt{PostTreeGraphRefiner}; $d_{\text{model}} = 192$, $4$ heads, $24$ distance RBFs, $<0.5\,\text{ms}$ per batched call) refines all $m_{\text{s2}}$ leaf point clouds $\{(\hat{Z}_i, \hat{\mathbf{r}}_i)\}_{i=1}^N$ via bounded zero-CoM displacements:
\begin{equation}
\label{eq:refiner}
\Delta \mathbf{r}_i^{(\ell)} = \frac{1}{\mathcal{N}_i} \sum_{j \in \mathcal{N}(i)} \psi_x\!\left(\mathbf{e}_{ij}^{(\ell)}\right) \frac{\mathbf{r}_i^{(\ell)} - \mathbf{r}_j^{(\ell)}}{\|\mathbf{r}_i^{(\ell)} - \mathbf{r}_j^{(\ell)}\| + \epsilon}, \qquad
\mathbf{r}_i^{\text{out}} = \hat{\mathbf{r}}_i + 0.25 \tanh\!\left(\frac{\tilde{\mathbf{r}}_i - \hat{\mathbf{r}}_i}{0.25}\right),
\end{equation}
paired with a symmetric bilinear covalent bond-order head $P(b_{ij} \in \{0, 1, 2, 3, \text{arom}\})$. \texttt{PostTree\allowbreak Graph\allowbreak Refiner} is distilled exclusively on dataset equilibrium conformers corrupted by random-$\text{SO}(3)$ octant-plane Morton shifts and voxel quantization noise, minimizing Kabsch-aligned ($\det \mathbf{R} = +1$) Huber coordinate error, 1--2/1--3 pairwise distance error, and bond-order cross-entropy.

\paragraph{Expected-\texttt{XYZ} Supervision, Selective CFG, and Valence Grounding.}
During rank-$64$ LoRA \citep{hu2022lora} adaptation (EMA decay $0.996$, keeping pretrained sub-voxel \texttt{XYZ} embeddings frozen on crystal-derived datasets), discrete $0.01\,\text{\AA}$ \texttt{XYZ} cross-entropy is augmented with a Smooth-$L_1$ loss over softmax-expected coordinates $\bar{\mathbf{r}}_i = \mathbf{o}_i + \delta_{xyz}\sum_b b\,p_i(b)$ across 1--2 covalent bonds ($d_{ij}^{\text{gt}} < 1.85\,\text{\AA}$) and 1--3 geminal pairs ($d_{ij}^{\text{gt}} < 2.65\,\text{\AA}$). At inference, tree-depth-annealed selective logit-CFG steers coarse tree and element heads while holding \texttt{XYZ} guidance strictly at $1.0$; guidance scale and affine input calibration parameters are selected using only the internal \texttt{TwoStage\allowbreak Octree\allowbreak ProcessCritic} on $128$ dev prompts. Pre-softmax coordination guards (\texttt{HeavyAtom\allowbreak ValenceHead}) and a zero-heavy-displacement \texttt{Valence\allowbreak Reconciler} ($\text{RMSD}_{\text{heavy}} \equiv 0.0000\,\text{\AA}$) resolve residual peripheral-H mismatches using the refiner's bond-order priors with zero MLIP calls.

\subsection{Batched GPU Decoding and Memory Scheduling}
\label{sec:engine}

To remove host-device synchronization stalls, Stage~1 uses masked in-place buffer writes that preserve native FlashAttention-2 across coarse steps while polling completion every $4$ steps. Stage~2 extracts placed-atom states via direct stride-$2$ slicing and compacts finished leaf rows every $6$ steps. Finally, KV caches allocate proportionally to target atom budget ($T_{\max} = \max(96, 3\lceil N_{\max}\rceil + 48 + 8(L_{\max}-6))$, with Stage-1 KV capped near $1.9 N_{\max} + 36$) alongside dynamic free-VRAM chunk scheduling.

\section{Experiments}
\label{sec:experiments}

We evaluate standalone single-output \textbf{MolGPAO} ($M_{\text{out}} \equiv 1$, zero external surrogates, zero RL, zero inference MLIP calls) across three benchmarks under $\text{SO}(3)$ random rotation ($p=0.80$, zero PCA canonicalization) and zero-$N_{\text{atoms}}$ property conditioning on \texttt{NVIDIA A800-80GB} GPUs: (1) $10{,}000$-molecule unconditional generation and $N \in \{30, 45, 60, 80, 100\}$ scaling on \texttt{GEOM-Drugs} \citep{axelrod2022geom} ($L_{\max}=7$, $14$ elements); (2) six-property conditional generation on the disjoint $50\text{k}/50\text{k}$ \texttt{QM9} benchmark \citep{ramakrishnan2014qm9,zeng2026propmolflow} ($L_{\max}=6$, $N=1{,}000$ test prompts per property, zero overlap with held-out \texttt{GVPRegressor} \citep{jing2021gvp}); and (3) pure ground-truth $\varepsilon_{\text{HOMO}}$ inverse design, $N \in \{20, 30, 40, 50, 60\}$ scaling, $6$-way mechanism ablation, and $42$-prompt live \texttt{GPU4PySCF PBE0/def2-TZVP/DF} \citep{sun2020pyscf,li2024gpu4pyscf} verification on \texttt{OE62} \citep{stuke2020oe62}.

\subsection{Unconditional Generation and Size Scaling on GEOM-Drugs}
\label{sec:exp_geom}

Starting from unconditional \texttt{Uni-3DAR} (\texttt{drug.pt}), \textbf{MolGPAO} requires $0.4914\,\text{GPU-h}$ of total post-training on \texttt{GEOM-Drugs}: $0.1479\,\text{h}$ of full-$\text{SO}(3)$ supervised adaptation, $0.0587\,\text{h}$ of \texttt{PostTreeGraphRefiner} distillation (lowering validation noisy bond-length MAE from $0.0668$ to $0.0126\,\text{\AA}$ at $99.27\%$ bond-order accuracy), and $0.2848\,\text{h}$ of \texttt{TwoStageOctreeProcessCritic} fitting ($99.26\%\text{--}99.39\%$ constraint-head accuracy).

\begin{figure}[!t]
  \centering
  \includegraphics[width=\textwidth]{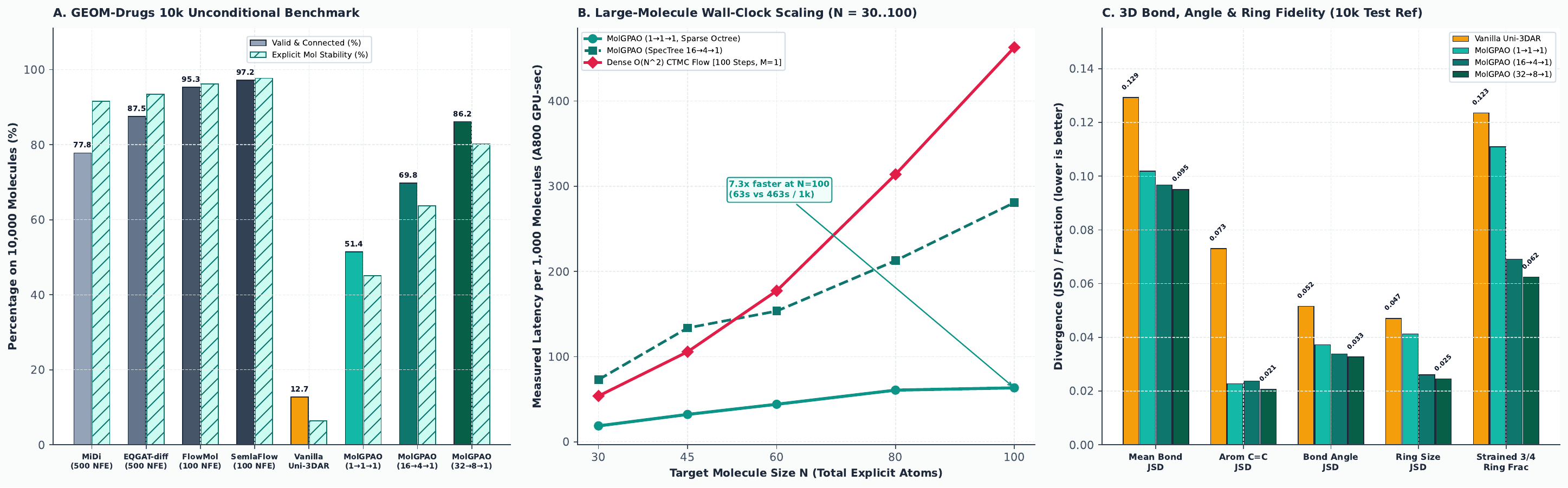}
  \caption{\textbf{Unconditional drug-scale generation ($10{,}000$ molecules) and $N=30\text{--}100$ scaling on \texttt{GEOM-Drugs} ($M_{\text{out}} \equiv 1$, zero inference MLIP).} \textbf{(A)} \texttt{Valid \& Connected} and explicit molecule stability across speculative search tiers. \textbf{(B)} Single-\texttt{A800} latency scaling across $N \in \{30,45,60,80,100\}$ vs.\ $100$-step \texttt{FlowMol}. \textbf{(C)} Bond-length, aromatic $\text{C=C}$, bond-angle, and ring-size JSD plus strained 3/4-ring fraction.}
  \label{fig:geom}
\end{figure}

Table~\ref{tab:geom}a and Figure~\ref{fig:geom}A,C compare single-output \textbf{MolGPAO} ($K \in \{1, 2, 4\}$) against \texttt{Vanilla Uni-3DAR}, coordinate-only diffusion (\texttt{EDM}, \texttt{GeoLDM}), and explicit-edge 2D+3D diffusion and flow-matching baselines (\texttt{MiDi-Adaptive}, \texttt{EQGAT-diff}, \texttt{FlowMol}, \texttt{SemlaFlow}) over $10{,}000$ generated conformers ($100.0\%$ unique and $100.0\%$ novel across all octree arms). \texttt{Vanilla Uni-3DAR} yields $12.70\%$ valid-and-connected structures and $6.42\%$ molecule stability due to cross-octant seam jumps and peripheral-H errors. Single-pass \textbf{MolGPAO} ($K=1$, $25.88\,\text{mol/s}$) lifts valid-and-connected yield $4.05\times$ to $51.41\%$. Speculative octree pruning at $K=4$ ($32\to 8\to 1$, $4.00\,\text{mol/s}$) raises \texttt{Valid \& Connected} $6.78\times$ to $86.15\%$---surpassing $500$-step explicit-edge \texttt{MiDi-Adaptive} ($77.8\%$) without dense $\mathcal{O}(N^2)$ bond matrices---while matching joint flow-matching atom valence stability ($99.83\%$) and lifting explicit molecule stability $12.5\times$ to $80.15\%$. Concurrently, aromatic $\text{C=C}$ bond-length Jensen--Shannon divergence (JSD) drops from $0.0730$ to $0.0207$, bond-angle JSD drops by $-36.6\%$, ring-size JSD drops by $-47.8\%$, strained 3/4-membered rings drop by $-49.4\%$, and $90$th-percentile zero-step \texttt{DPA-4} force falls from $17.34$ to $13.92\,\text{eV/\AA}$.

Table~\ref{tab:geom}b and Figure~\ref{fig:geom}B test $\mathcal{O}(N \log N)$ octree sparsity against $\mathcal{O}(N^2)$ $100$-step \texttt{FlowMol} across exact budgets $N \in \{30, 45, 60, 80, 100\}$ ($250$ molecules per bucket) on identical \texttt{A800} hardware. From $N=30$ to $N=100$ explicit atoms, \texttt{FlowMol} slows $8.6\times$ ($53.8 \to 463.1\,\text{s/1k}$) while valid-connected yield falls from $97.6\%$ to $14.0\%$. By contrast, \textbf{MolGPAO} ($1\to 1\to 1$) reaches $63.4\,\text{s/1k}$ at $N=100$ ($7.3\times$ faster than \texttt{FlowMol}) with $22.8\%$ valid-connected yield, while \textbf{MolGPAO} ($16\to 4\to 1$, $280.8\,\text{s/1k}$) is simultaneously $1.65\times$ faster and $1.97\times$ higher in valid-connected yield ($27.6\%$ vs.\ $14.0\%$, $0.1157$ vs.\ $0.4038$ bond JSD).

\begin{table}[!t]
  \centering
  \caption{Unconditional $10{,}000$-molecule generation and controlled atom-count scaling on \texttt{GEOM-Drugs} ($M_{\text{out}} \equiv 1$, single \texttt{NVIDIA A800-80GB}).}
  \label{tab:geom}
  \scriptsize
  \setlength{\tabcolsep}{3.0pt}
  \textbf{a}, Standard $10{,}000$-molecule \texttt{GEOM-Drugs} benchmark ($14$ elements).\\[2pt]
  \resizebox{\textwidth}{!}{%
  \begin{tabular}{@{}llrrrrrrrrrr@{}}
    \toprule
    Method & Paradigm (Steps / Internal Funnel) & \shortstack[r]{Speed\\(mol/s)} & \shortstack[r]{Valid \&\\Conn.\ (\%)} & \shortstack[r]{Atom\\Stab.\ (\%)} & \shortstack[r]{Mol\\Stab.\ (\%)} & \shortstack[r]{Bond\\JSD $\downarrow$} & \shortstack[r]{Arom C=C\\JSD $\downarrow$} & \shortstack[r]{Angle\\JSD $\downarrow$} & \shortstack[r]{Ring\\JSD $\downarrow$} & \shortstack[r]{3/4-Ring\\(\%) $\downarrow$} & \shortstack[r]{$F_{\max}$ Med/P90\\($\text{eV/\AA}$)} \\
    \midrule
    EDM \citep{hoogeboom2022edm} & Coord.\ $E(3)$ Diff.\ (1000) & --- & --- & 81.3\rlap{$^*$} & 0.2\rlap{$^*$} & --- & --- & --- & --- & --- & --- \\
    GeoLDM \citep{xu2023geoldm} & Latent $E(3)$ Diff.\ (1000) & --- & --- & 84.4\rlap{$^*$} & 5.2\rlap{$^*$} & --- & --- & --- & --- & --- & --- \\
    MiDi-Adaptive \citep{vignac2023midi} & Joint 2D+3D Diff.\ (500) & --- & 77.8 & 99.8 & 91.6 & --- & --- & --- & --- & --- & --- \\
    EQGAT-diff \citep{le2023eqgatdiff} & Joint Equiv.\ Diff.\ (500) & --- & 87.5 & 99.8 & 93.4 & --- & --- & --- & --- & --- & --- \\
    FlowMol \citep{dunn2024flowmol} & Joint CTMC Flow (100) & --- & 95.3 & 99.7 & 96.2 & --- & --- & --- & --- & --- & --- \\
    SemlaFlow \citep{irwin2025semlaflow} & Joint Equiv.\ OT Flow (100) & --- & \textbf{97.2} & \textbf{99.8} & \textbf{97.7} & --- & --- & --- & --- & --- & --- \\
    \midrule
    Vanilla Uni-3DAR (\texttt{drug.pt}) & Unbonded Octree ($1\to 1\to 1$) & \textbf{59.12} & 12.70 & 97.63 & 6.42 & 0.1293 & 0.0730 & 0.0516 & 0.0471 & 12.35 & 8.03 / 17.34 \\
    \textbf{MolGPAO} ($K=1$) & Octree + Refiner ($1\to 1\to 1$) & 25.88 & 51.41 & 99.69 & 45.14 & 0.1019 & 0.0227 & 0.0373 & 0.0412 & 11.09 & 7.49 / 17.43 \\
    \textbf{MolGPAO} ($K=2$) & SpecTree + Refiner ($16\to 4\to 1$) & 7.67 & 69.78 & 99.78 & 63.67 & 0.0968 & 0.0238 & 0.0338 & 0.0260 & 6.91 & \textbf{6.89} / 14.52 \\
    \textbf{MolGPAO} ($K=4$) & SpecTree + Refiner ($32\to 8\to 1$) & 4.00 & \textbf{86.15} & \textbf{99.83} & \textbf{80.15} & \textbf{0.0951} & \textbf{0.0207} & \textbf{0.0327} & \textbf{0.0246} & \textbf{6.25} & 6.93 / \textbf{13.92} \\
    \bottomrule
  \end{tabular}%
  }

  \vspace{5pt}
  \textbf{b}, Controlled scaling across exact target atom budgets $N \in \{30, 45, 60, 80, 100\}$ ($250$ molecules per bucket).\\[2pt]
  \resizebox{\textwidth}{!}{%
  \begin{tabular}{@{}lrrrrrrrrrr@{}}
    \toprule
    & \multicolumn{2}{c}{$N=30$} & \multicolumn{2}{c}{$N=45$} & \multicolumn{2}{c}{$N=60$} & \multicolumn{2}{c}{$N=80$} & \multicolumn{2}{c}{$N=100$} \\
    \cmidrule(lr){2-3}\cmidrule(lr){4-5}\cmidrule(lr){6-7}\cmidrule(lr){8-9}\cmidrule(lr){10-11}
    Arm (Identical A800 Hardware) & s/1k $\downarrow$ & ValidConn & s/1k $\downarrow$ & ValidConn & s/1k $\downarrow$ & ValidConn & s/1k $\downarrow$ & ValidConn & s/1k $\downarrow$ & ValidConn \\
    \midrule
    FlowMol Dense $\mathcal{O}(N^2)$ ODE-100 ($M_{\text{out}}=1$) & 53.8 & \textbf{97.6\%} & 105.7 & \textbf{89.6\%} & 177.3 & \textbf{67.2\%} & 313.9 & \textbf{38.8\%} & 463.1 & 14.0\% \\
    \textbf{MolGPAO} ($K=1$: $1\to 1\to 1$) & \textbf{18.7} & 62.0\% & \textbf{32.1} & 34.0\% & \textbf{44.1} & 26.4\% & \textbf{60.7} & 27.6\% & \textbf{63.4} & 22.8\% \\
    \textbf{MolGPAO} ($K=2$: $16\to 4\to 1$) & 73.0 & \textbf{97.6\%} & 133.7 & 76.4\% & 153.5 & 34.8\% & 212.7 & 32.8\% & 280.8 & \textbf{27.6\%} \\
    \bottomrule
  \end{tabular}%
  }
  \vspace{2pt}\par{\scriptsize All octree arms achieve $100.0\%$ uniqueness and novelty on \texttt{GEOM-Drugs}. $^*$EDM and GeoLDM report distance-lookup stability without explicit bond perception; joint 2D+3D baselines and octree rows report RDKit valence stability on perceived molecular graphs.}
\end{table}

\subsection{Property-Conditional Generation on QM9}
\label{sec:exp_qm9}

We trained a single unified $6$-property full-$\text{SO}(3)$ \textbf{MolGPAO} model on the disjoint $50\text{k}/50\text{k}$ \texttt{PropMolFlow} \texttt{QM9} split \citep{ramakrishnan2014qm9,zeng2026propmolflow} ($44{,}453$ train and $1{,}500$ dev conformers, zero overlap with the held-out \texttt{GVPRegressor} evaluator \citep{jing2021gvp}, $N=1{,}000$ test prompts per property). Whereas \texttt{PropMolFlow} trains six property-specific checkpoints in $155.97\,\text{GPU-h}$ with explicit $(t, N_{\text{atoms}})$ conditioning, \textbf{MolGPAO} trains one unified $6$-property checkpoint in $0.4303\,\text{GPU-h}$ ($362\times$ less compute: $0.290\,\text{h}$ SFT, $0.025\,\text{h}$ \texttt{PostTreeGraphRefiner}, and $0.115\,\text{h}$ ground-truth \texttt{TwoStageOctreeProcessCritic}) with zero PCA canonicalization, zero $N_{\text{atoms}}$ input to the property conditioner, and zero external surrogates.

Table~\ref{tab:qm9}, Figure~\ref{fig:qm9_oe62}A, and Figure~\ref{fig:ablation}C highlight three results. First, supervised solely on ground-truth \texttt{QM9} labels with contrast-group decoupling, the $64$-RBF multipole critic reaches blind mismatched-target dev Pearson correlations of $r = (0.9983, 0.9940, 0.9867, 0.9959, 0.9950, 0.9989)$ across $(\alpha, \Delta\varepsilon, \varepsilon_{\text{HOMO}}, \varepsilon_{\text{LUMO}}, \mu, C_v)$. Second, synchronization-free GPU decoding accelerates $K=1$ output to $127.61\,\text{mol/s}$ ($4.4\times$ faster than \texttt{PropMolFlow}'s $28.73\,\text{mol/s}$), while speculative funnels lift chemical legality from $50.1\%$ (\texttt{Vanilla Uni-3DAR}) to $99.97\%\text{--}100.0\%$ across $K \ge 2$ and tighten atom-count error $|\Delta N|$ from $2.56\text{--}3.07$ to $0.22\text{--}0.38$ atoms at $K=16$. Third, with zero external oracle calls, $K=4$ ($16.51\,\text{mol/s}$) beats $6$-model \texttt{EDM} and \texttt{GeoLDM} on all four intensive electronic/polar properties ($\Delta\varepsilon, \varepsilon_{\text{HOMO}}, \varepsilon_{\text{LUMO}}, \mu$), while $K=16$ ($4.47\,\text{mol/s}$) surpasses $156\,\text{GPU-h}$ \texttt{PropMolFlow} on $\varepsilon_{\text{HOMO}}$ ($221.84$ vs.\ $253.99\,\text{meV}$, $-12.7\%$) and $\mu$ ($0.4902$ vs.\ $0.6202\,\text{D}$, $-21.0\%$).

\begin{table}[!t]
  \centering
  \caption{Standalone six-property conditional generation on the disjoint $50\text{k}/50\text{k}$ \texttt{QM9} split ($N=1{,}000$ test prompts per property, $M_{\text{out}} \equiv 1$, zero external surrogates, single \texttt{NVIDIA A800-80GB}).}
  \label{tab:qm9}
  \scriptsize
  \setlength{\tabcolsep}{2.8pt}
  \resizebox{\textwidth}{!}{%
  \begin{tabular}{@{}lccrrrrrrrrr@{}}
    \toprule
    Method / Internal Speculative Funnel & \# & $M_{\text{out}}$ & \shortstack[r]{$\alpha$\\($\text{Bohr}^3$)} & \shortstack[r]{$\Delta\varepsilon$\\(meV)} & \shortstack[r]{$\varepsilon_{\text{HOMO}}$\\(meV)} & \shortstack[r]{$\varepsilon_{\text{LUMO}}$\\(meV)} & \shortstack[r]{$\mu$\\(D)} & \shortstack[r]{$C_v$\\($\text{cal/mol}\cdot\text{K}$)} & \shortstack[r]{Legal (\%) /\\$|\Delta N|$ (atoms)} & \shortstack[r]{Train\\(GPU-h)} & \shortstack[r]{Out\\(mol/s)} \\
    \midrule
    GVP Oracle Holdout Floor ($n=973$) & --- & --- & 0.5544 & 190.08 & 83.78 & 135.89 & 0.1422 & 0.1726 & 97.3 / --- & --- & --- \\
    EDM (published) \citep{hoogeboom2022edm} & 6 & 1 & 2.76 & 655.0 & 356.0 & 584.0 & 1.111 & 1.101 & --- & --- & --- \\
    GeoLDM (published) \citep{xu2023geoldm} & 6 & 1 & 2.37 & 587.0 & 340.0 & 522.0 & 1.108 & 1.025 & --- & --- & --- \\
    PropMolFlow ($[t,N]$, published) \citep{zeng2026propmolflow} & 6 & 1 & \textbf{1.3115} & \textbf{390.98} & 253.99 & \textbf{314.52} & 0.6202 & \textbf{0.6259} & 95.0--98.9 / $\equiv$0 & 155.97 & 28.73 \\
    \midrule
    Vanilla Uni-3DAR ($1\times 6$-prop SFT, $1\to 1\to 1$) & 1 & 1 & 7.7991 & 1034.61 & 602.19 & 1115.91 & 1.7876 & 3.3365 & 50.1 / 2.56--3.07 & \textbf{0.23} & 97.27 \\
    \textbf{MolGPAO} ($K=1$: $1\to 1\to 1$) & 1 & 1 & 5.7912 & 1141.13 & 569.92 & 1076.80 & 1.7685 & 3.0132 & 83.0 / 0.83--0.99 & 0.43 & \textbf{127.61} \\
    \textbf{MolGPAO} ($K=2$: $10\to 4\to 1$) & 1 & 1 & 3.6310 & 634.26 & 304.95 & 605.93 & 0.8818 & 2.0830 & 99.97 / 0.46--0.61 & 0.43 & 28.04 \\
    \textbf{MolGPAO} ($K=4$: $20\to 6\to 1$) & 1 & 1 & 3.5413 & 552.76 & 247.32 & 501.84 & 0.7065 & 1.8279 & \textbf{100.0} / 0.38--0.51 & 0.43 & 16.51 \\
    \textbf{MolGPAO} ($K=8$: $38\to 12\to 1$) & 1 & 1 & 3.2636 & 464.97 & 233.76 & 419.78 & 0.5680 & 1.6988 & 99.98 / 0.27--0.39 & 0.43 & 8.72 \\
    \textbf{MolGPAO} ($K=16$: $74\to 22\to 1$) & 1 & 1 & 3.1769 & 406.34 & \textbf{221.84} & 385.35 & \textbf{0.4902} & 1.5828 & \textbf{100.0} / \textbf{0.22--0.38} & 0.43 & 4.47 \\
    \bottomrule
  \end{tabular}%
  }
  \vspace{2pt}\par{\scriptsize Every row emits $M_{\text{out}} = 1$ molecule per prompt with zero overlap against the held-out \texttt{GVPRegressor} evaluator. At $K=16$, \textbf{MolGPAO} records test Pearson $r = (0.9229, 0.8813, 0.8744, 0.8992, 0.8302, 0.9148)$ with $86.0\%\text{--}90.6\%$ novelty and $99.7\%\text{--}100.0\%$ uniqueness.}
\end{table}

\subsection{Orbital Design and Quantum Verification on OE62}
\label{sec:exp_oe62}

Finally, we evaluated standalone \textbf{MolGPAO} on scaffold-disjoint \texttt{OE62} organic semiconductors \citep{stuke2020oe62,zhao2025virtual,wang2026mimflow,zhao2026o2gen} conditioned solely on pure \texttt{PBE0} ground-truth $\varepsilon_{\text{HOMO}}$ labels ($22{,}901$ train and $3{,}188$ validation equilibrium gas-phase DFT conformers; zero synthetic accessibility filters, zero external surrogates, zero RL, zero inference MLIP calls). Total post-training compute from \texttt{drug.pt} is $0.5604\,\text{GPU-h}$ ($0.3949\,\text{h}$ rank-$64$ LoRA SFT, $0.0254\,\text{h}$ synthetic-seam \texttt{PostTreeGraphRefiner} achieving $0.0143\,\text{\AA}$ bond MAE and $98.61\%$ bond-order accuracy, and $0.1401\,\text{h}$ ground-truth \texttt{TwoStageOctreeProcessCritic}). We compare against single-shot \texttt{PropMolFlow} ($M_{\text{out}}=1$) trained from scratch across $3.0$, $12.0$, and $52.0\,\text{GPU-h}$ on identical $N=1{,}792$ fixed $(t_i, N_{\text{atoms},i})$ prompts (mean $30.38$ atoms) using a blinded DFT-calibrated \texttt{DPA-4} orbital/force reference ($|\hat{y}_{\text{HOMO}} - t_i| \le 0.25\,\text{eV}$, clean closed-shell valence) and $42$ deterministic prompts under live \texttt{GPU4PySCF PBE0/def2-TZVP/DF}.

\begin{figure}[!t]
  \centering
  \includegraphics[width=\textwidth]{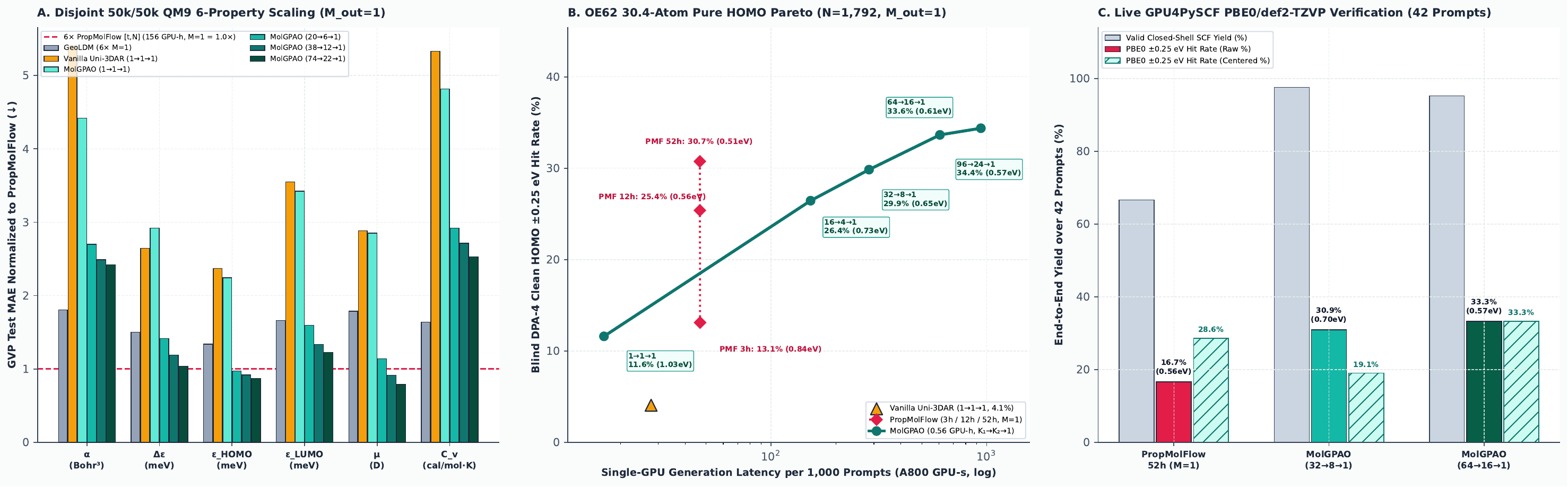}
  \caption{\textbf{Standalone conditional scaling ($M_{\text{out}} \equiv 1$, zero external surrogates) on \texttt{QM9} and \texttt{OE62} with live \texttt{PBE0/def2-TZVP} verification.} \textbf{(A)} Disjoint \texttt{QM9} test \texttt{GVPRegressor} MAE normalized to $6$-model \texttt{PropMolFlow}. \textbf{(B)} \texttt{OE62} latency vs.\ blind \texttt{DPA-4} clean hit Pareto ($N=1{,}792$). \textbf{(C)} Unbiased $42$-prompt live \texttt{GPU4PySCF PBE0/def2-TZVP} verification.}
  \label{fig:qm9_oe62}
\end{figure}

\paragraph{Speculative Scaling and Live \texttt{PBE0} Verification.}
In Table~\ref{tab:oe62}a and Figure~\ref{fig:qm9_oe62}B, raw SFT \texttt{Vanilla Uni-3DAR} suffers $32.39\%$ exotic charges, $|\Delta N| = 10.03$ atoms, and $4.07\%$ blind clean hit rate. \textbf{MolGPAO} ($0.56\,\text{GPU-h}$) eliminates exotic charges ($0.00\%$ across all tiers), accelerates $K=1$ throughput to $59.59\,\text{mol/s}$, and scales monotonically through $K=2$ ($26.45\%$ clean hits, surpassing $12.0\,\text{GPU-h}$ \texttt{PropMolFlow} at $25.39\%$), $K=8$ ($33.65\%$, $1{,}439$ unique Murcko scaffolds), and $K=12$ ($34.38\%$, $616/1{,}792$, $100.0\%$ legality, $|\Delta N|=0.74$ atoms, slope $0.99$), beating $52.00\,\text{GPU-h}$ \texttt{PropMolFlow} ($30.75\%$ clean hits, $6.09\%$ exotic charges) by $+3.63\,\text{pp}$ at $92.8\times$ lower training compute. On $42$-prompt live \texttt{GPU4PySCF PBE0/def2-TZVP/DF} verification (Figure~\ref{fig:qm9_oe62}C), $52\,\text{GPU-h}$ \texttt{PropMolFlow} yields $28/42$ ($66.7\%$) closed-shell SCF-converged structures and $16.67\%$ ($7/42$) raw \texttt{PBE0} hits, whereas $0.56\,\text{GPU-h}$ \textbf{MolGPAO} reaches $97.6\%$ ($K=4$) and $95.2\%$ ($K=8$) closed-shell SCF yield with $33.33\%$ ($14/42$, $2.0\times$ \texttt{PropMolFlow}) end-to-end \texttt{PBE0} clean hit rate.

\begin{figure}[!t]
  \centering
  \includegraphics[width=\textwidth]{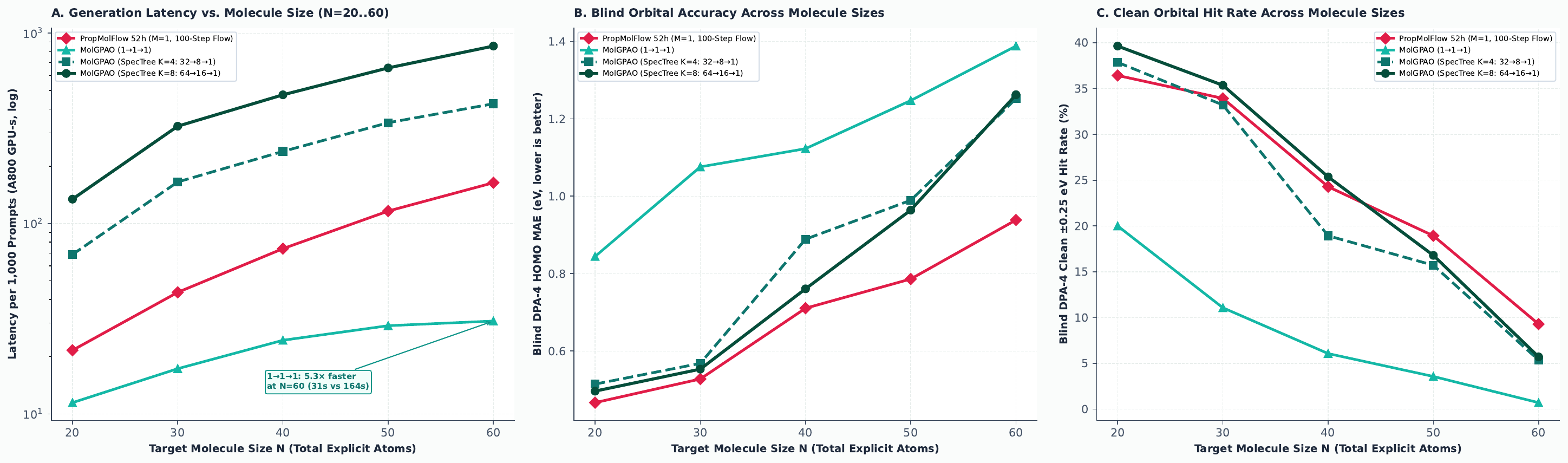}
  \caption{\textbf{Controlled molecule-size scaling across exact atom budgets $N \in \{20, 30, 40, 50, 60\}$ on \texttt{OE62} ($280$ fixed-size prompts per bucket, $M_{\text{out}} \equiv 1$).} \textbf{(A)} Single-GPU latency per $1{,}000$ prompts. \textbf{(B)} Blind \texttt{DPA-4} \texttt{HOMO} MAE vs.\ $N$. \textbf{(C)} Blind clean $\pm 0.25\,\text{eV}$ hit rate vs.\ $N$.}
  \label{fig:oe62_atoms}
\end{figure}

\begin{figure}[!t]
  \centering
  \includegraphics[width=\textwidth]{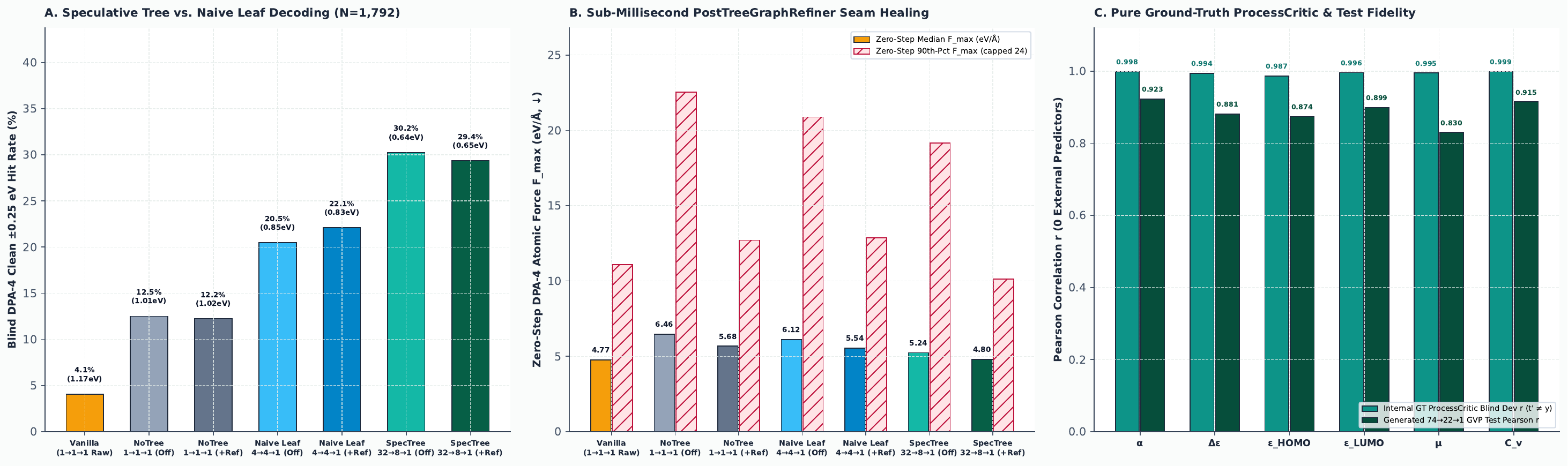}
  \caption{\textbf{Internal mechanism ablation on \texttt{OE62} ($N=1{,}792$, $M_{\text{out}} \equiv 1$) and anti-shortcut critic correlation.} \textbf{(A)} Two-stage speculative octree pruning ($32\to 8\to 1$) vs.\ leaf-complete ($4\to 4\to 1$) and single-pass ($1\to 1\to 1$) decoding. \textbf{(B)} Zero-step $90$th-percentile \texttt{DPA-4} atomic force reduction from sub-millisecond \texttt{PostTreeGraphRefiner} seam healing. \textbf{(C)} Blind mismatched-target dev Pearson $r$ of \texttt{OctreeElementProcessCritic} and test \texttt{GVPRegressor} Pearson $r$ across six \texttt{QM9} properties.}
  \label{fig:ablation}
\end{figure}

\paragraph{Atom-Size Scaling and Internal Mechanism Ablation.}
Across exact atom budgets $N \in \{20, 30, 40, 50, 60\}$ ($280$ prompts per bucket; Figure~\ref{fig:oe62_atoms}), $0.56\,\text{GPU-h}$ \textbf{MolGPAO} ($K=8$) leads $52\,\text{GPU-h}$ \texttt{PropMolFlow} in clean hit rate at $N=20$ ($39.64\%$ vs.\ $36.43\%$), $N=30$ ($35.36\%$ vs.\ $33.93\%$), and $N=40$ ($25.36\%$ vs.\ $24.29\%$), while running $5.3\times$ faster at $N=60$ ($30.8$ vs.\ $163.9\,\text{s/1k}$ at $K=1$) and maintaining $>99\%$ legality at $K \ge 4$ (vs.\ $67.5\%$ for \texttt{PropMolFlow}). The $6$-way internal ablation across all $N=1{,}792$ prompts (Table~\ref{tab:oe62}b and Figure~\ref{fig:ablation}A,B) confirms two orthogonal contributions: two-stage speculative octree pruning ($32\to 8\to 1$ vs.\ leaf-complete $4\to 4\to 1$) lifts clean hit rate by $+7.25\text{--}9.77\,\text{pp}$ while halving $|\Delta N|$ ($1.94 \to 0.98$ atoms), and $<0.5\,\text{ms}$ \texttt{PostTreeGraphRefiner} slashes $90$th-percentile zero-step $F_{\max}$ by $38\%\text{--}47\%$ ($19.16 \to 10.12\,\text{eV/\AA}$ at $32\to 8\to 1$) without MLIP relaxation.

\begin{table}[!t]
  \centering
  \caption{Standalone \texttt{OE62 HOMO} inverse design benchmark ($N=1{,}792$ prompts, mean $30.4$ atoms, $M_{\text{out}} \equiv 1$, zero external surrogates, zero inference MLIP) and $6$-way internal mechanism ablation.}
  \label{tab:oe62}
  \scriptsize
  \setlength{\tabcolsep}{2.3pt}
  \textbf{a}, Cross-paradigm single-output scaling and $42$-prompt live \texttt{GPU4PySCF PBE0/def2-TZVP} verification.\\[2pt]
  \resizebox{\textwidth}{!}{%
  \begin{tabular}{@{}llrrrrrrrrrr@{}}
    \toprule
    Method / Paradigm ($M_{\text{out}} \equiv 1$) & \shortstack[l]{Internal\\Funnel} & \shortstack[r]{Train\\(GPU-h)} & \shortstack[r]{Out\\(mol/s)} & \shortstack[r]{Atoms /\\$|\Delta N|$} & \shortstack[r]{Legal /\\Exotic $Q$ (\%)} & \shortstack[r]{Unique\\Scaffolds} & \shortstack[r]{$F_{\max}$ Med/P90\\($\text{eV/\AA}$)} & \shortstack[r]{Blind Clean\\Hits (\%)} & \shortstack[r]{Blind MAE /\\Slope} & \shortstack[r]{PBE0 SCF\\Yield (\%)} & \shortstack[r]{Live PBE0 Hit\\(Raw / Cen)} \\
    \midrule
    PropMolFlow (100-Step Flow) \citep{zeng2026propmolflow} & $1\to 1$ & 3.00 & 21.14 & 29.7 / \textbf{0.00} & 69.4 / 2.25 & 984 & 4.68 / 13.44 & 235 (13.11\%) & 0.843 / 0.64 & --- & --- \\
    PropMolFlow (100-Step Flow) \citep{zeng2026propmolflow} & $1\to 1$ & 12.00 & 21.20 & 30.2 / \textbf{0.00} & 84.5 / 3.70 & 1232 & 3.39 / 5.82 & 455 (25.39\%) & 0.563 / 0.84 & --- & --- \\
    PropMolFlow (100-Step Flow) \citep{zeng2026propmolflow} & $1\to 1$ & 52.00 & 21.20 & 30.1 / \textbf{0.00} & 89.8 / 6.09 & 1221 & \textbf{3.09} / \textbf{5.08} & 551 (30.75\%) & \textbf{0.512} / 0.87 & 66.7 & 16.7\% / 28.6\% \\
    \midrule
    Vanilla Uni-3DAR (Raw SFT) & $1\to 1\to 1$ & \textbf{0.25} & 36.02 & 30.4 / 10.03 & 33.4 / 32.39 & 505 & 4.77 / 11.10 & 73 (4.07\%) & 1.173 / 1.55 & --- & --- \\
    \textbf{MolGPAO} ($K=1$) & $1\to 1\to 1$ & 0.56 & \textbf{59.59} & 28.9 / 1.93 & 70.3 / \textbf{0.00} & 1046 & 5.88 / 14.67 & 208 (11.61\%) & 1.029 / 0.89 & --- & --- \\
    \textbf{MolGPAO} ($K=2$) & $16\to 4\to 1$ & 0.56 & 6.56 & 30.2 / 1.14 & 99.1 / \textbf{0.00} & 1410 & 4.99 / 11.01 & 474 (26.45\%) & 0.730 / 0.92 & --- & --- \\
    \textbf{MolGPAO} ($K=4$) & $32\to 8\to 1$ & 0.56 & 3.51 & 30.3 / 0.98 & 99.7 / \textbf{0.00} & 1424 & 4.89 / 10.62 & 535 (29.85\%) & 0.653 / 0.93 & \textbf{97.6} & 31.0\% / 19.0\% \\
    \textbf{MolGPAO} ($K=8$) & $64\to 16\to 1$ & 0.56 & 1.65 & 30.5 / 0.81 & 99.8 / \textbf{0.00} & \textbf{1439} & 4.72 / 9.90 & 603 (33.65\%) & 0.606 / 0.97 & 95.2 & \textbf{33.3\%} / \textbf{33.3\%} \\
    \textbf{MolGPAO} ($K=12$) & $96\to 24\to 1$ & 0.56 & 1.06 & 30.5 / 0.74 & \textbf{100.0} / \textbf{0.00} & 1422 & 4.54 / 9.13 & \textbf{616} (\textbf{34.38\%}) & 0.567 / \textbf{0.99} & --- & --- \\
    \bottomrule
  \end{tabular}%
  }

  \vspace{5pt}
  \textbf{b}, Internal speculative decoding and seam-healing ablation ($N=1{,}792$ prompts, $M_{\text{out}} \equiv 1$).\\[2pt]
  \resizebox{\textwidth}{!}{%
  \begin{tabular}{@{}lccrrrrrrrr@{}}
    \toprule
    Ablation Configuration ($M_{\text{out}} \equiv 1$) & Internal Funnel & \texttt{PostTreeGraphRefiner} & \shortstack[r]{Out\\(mol/s)} & \shortstack[r]{Legal\\(\%)} & \shortstack[r]{$|\Delta N|$\\(atoms)} & \shortstack[r]{$F_{\max}$ Med\\($\text{eV/\AA}$)} & \shortstack[r]{$F_{\max}$ P90\\($\text{eV/\AA}$)} & \shortstack[r]{Blind Clean\\Hits (\%)} & \shortstack[r]{Blind MAE\\(eV)} & \shortstack[r]{Slope /\\Pearson $r$} \\
    \midrule
    Single-Pass Octree (No Critic) & $1\to 1\to 1$ & Off & \textbf{59.63} & 73.5 & 1.96 & 6.46 & 22.54 & 224 (12.50\%) & 1.006 & 0.93 / 0.44 \\
    Single-Pass Octree + Seam Refiner & $1\to 1\to 1$ & On ($<0.5\,\text{ms}$) & 58.96 & 71.8 & 1.95 & 5.68 & 12.70 & 219 (12.22\%) & 1.024 & 0.97 / 0.46 \\
    Naive Leaf-Complete Best-of-$4$ & $4\to 4\to 1$ & Off & 19.20 & 98.9 & 1.60 & 6.12 & 20.88 & 367 (20.48\%) & 0.846 & 0.89 / 0.45 \\
    Naive Leaf-Complete + Seam Refiner & $4\to 4\to 1$ & On ($<0.5\,\text{ms}$) & 18.72 & 98.6 & 1.94 & 5.55 & 12.86 & 396 (22.10\%) & 0.828 & 0.89 / 0.46 \\
    Speculative Two-Stage Octree Only & $32\to 8\to 1$ & Off & 3.30 & \textbf{99.9} & 1.05 & 5.24 & 19.16 & \textbf{542} (\textbf{30.25\%}) & \textbf{0.641} & \textbf{0.95} / \textbf{0.52} \\
    \textbf{Full MolGPAO} & $32\to 8\to 1$ & On ($<0.5\,\text{ms}$) & 3.13 & 99.7 & \textbf{0.98} & \textbf{4.81} & \textbf{10.12} & 526 (29.35\%) & 0.650 & \textbf{0.95} / \textbf{0.52} \\
    \bottomrule
  \end{tabular}%
  }
  \vspace{2pt}\par{\scriptsize Blind clean hits require $|\hat{y}_{\text{HOMO}} - t_i| \le 0.25\,\text{eV}$ under the isolated \texttt{DPA-4} reference ($\text{MAE} = 0.1578\,\text{eV}$, $r=0.9542$ against \texttt{PBE0/def2-TZVP}) with closed-shell neutral valence.}
\end{table}

\section{Conclusion}
\label{sec:conclusion}

We have presented \textbf{MolGPAO}, transforming unbonded 3D octree autoregression into an equilibrium-accurate, valence-clean, self-contained 3D molecular generation engine through seamless speculative decoding. Operating strictly with single-molecule output ($M_{\text{out}} \equiv 1$) and zero external surrogates, zero reinforcement learning, and zero inference-time MLIP relaxation, \textbf{MolGPAO} unites scale-free budget modulation, pause-and-gather speculative tree pruning, an anti-shortcut equivariant multipole process critic, sub-millisecond $SE(3)$ seam healing, and synchronization-free GPU leaf decoding, delivering $\mathcal{O}(N \log N)$ scaling and outperforming $52\text{--}156\,\text{GPU-h}$ flow-matching baselines across \texttt{GEOM-Drugs}, \texttt{QM9}, and \texttt{PBE0}-verified \texttt{OE62} at two orders of magnitude lower training compute.

\bibliographystyle{unsrtnat}
\bibliography{references}

\end{document}